\pdfoutput=1

\documentclass[11pt,table]{article}
\usepackage{newtxtext}
\usepackage{float}
\usepackage[preprint]{acl}

\usepackage{times}
\usepackage{latexsym}
\usepackage{graphicx}
\usepackage{amsmath}
\usepackage{pdfpages}
\usepackage{colortbl}
\usepackage{booktabs}
\usepackage{multirow}

\usepackage[T1]{fontenc}

\usepackage[utf8]{inputenc}

\usepackage{microtype}

\usepackage{inconsolata}
\usepackage{multirow}

\usepackage{xspace}

\newcommand{\tabx}{\textsc{TabXEval}\xspace}
\newcommand{\tabbench}{\textsc{TabXBench}\xspace}

\title{	\tabx{}: Why this is a Bad Table?  An eXhaustive Rubric for \\ Table Evaluation}

\author{%
  \thanks{These authors contributed equally to this work.}\textbf{Vihang Pancholi\raisebox{0.75ex}{\includegraphics[height=2ex]{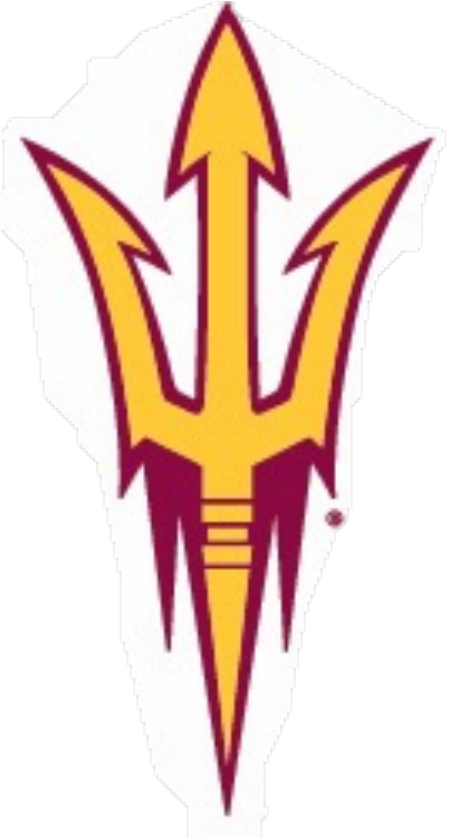}}} \quad
  \footnotemark[1]\textbf{Jainit Bafna\raisebox{1ex}{\includegraphics[height=2ex]{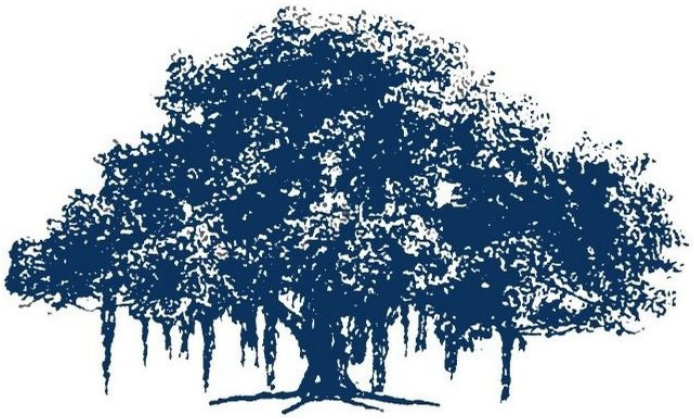}}} \quad
  \footnotemark[1]\textbf{Tejas Anvekar\raisebox{0.75ex}{\includegraphics[height=2ex]{asu_logo.pdf}}} \\
  \textbf{Manish Shrivastava\raisebox{1ex}{\includegraphics[height=2ex]{iiith_logo.pdf}}} \quad
  \thanks{Primary supervisor and corresponding author of this work.}\textbf{Vivek Gupta\raisebox{0.75ex}{\includegraphics[height=2ex]{asu_logo.pdf}}} \\[0.6em]
  \raisebox{0.75ex}{\includegraphics[height=2ex]{asu_logo.pdf}}\hspace{0.3ex}Arizona State University \quad
  \raisebox{1ex}{\includegraphics[height=2ex]{iiith_logo.pdf}}\hspace{0.3ex}IIIT Hyderabad \\
  \texttt{\{vpancho1,tanvekar,vgupt140\}@asu.edu} \\
  \texttt{\{jainit.bafna,m.shrivastava\}@research.iiit.ac.in}
}

\begin{document}


\maketitle

\begin{abstract}
Evaluating tables qualitatively and quantitatively poses a significant challenge, as standard metrics often overlook subtle structural and content-level discrepancies. To address this, we propose a rubric-based evaluation framework that integrates multi-level structural descriptors with fine-grained contextual signals, enabling more precise and consistent table comparison. Building on this, we introduce \textbf{\tabx{}}, an eXhaustive and eXplainable two-phase evaluation framework. \tabx{} first aligns reference and predicted tables structurally via \textsc{TabAlign}, then performs semantic and syntactic comparison using \textsc{TabCompare}, offering interpretable and granular feedback. We evaluate \tabx{} on \textbf{\tabbench{}}, a diverse, multi-domain benchmark featuring realistic table perturbations and human annotations. A sensitivity-specificity analysis further demonstrates the robustness and explainability of \tabx{} across varied table tasks. Code and data are available at~\href{https://coral-lab-asu.github.io/tabxeval/}{https://coral-lab-asu.github.io/tabxeval/}.

\end{abstract}
\begin{figure}[t]
    \centering
    \includegraphics[width=\linewidth]{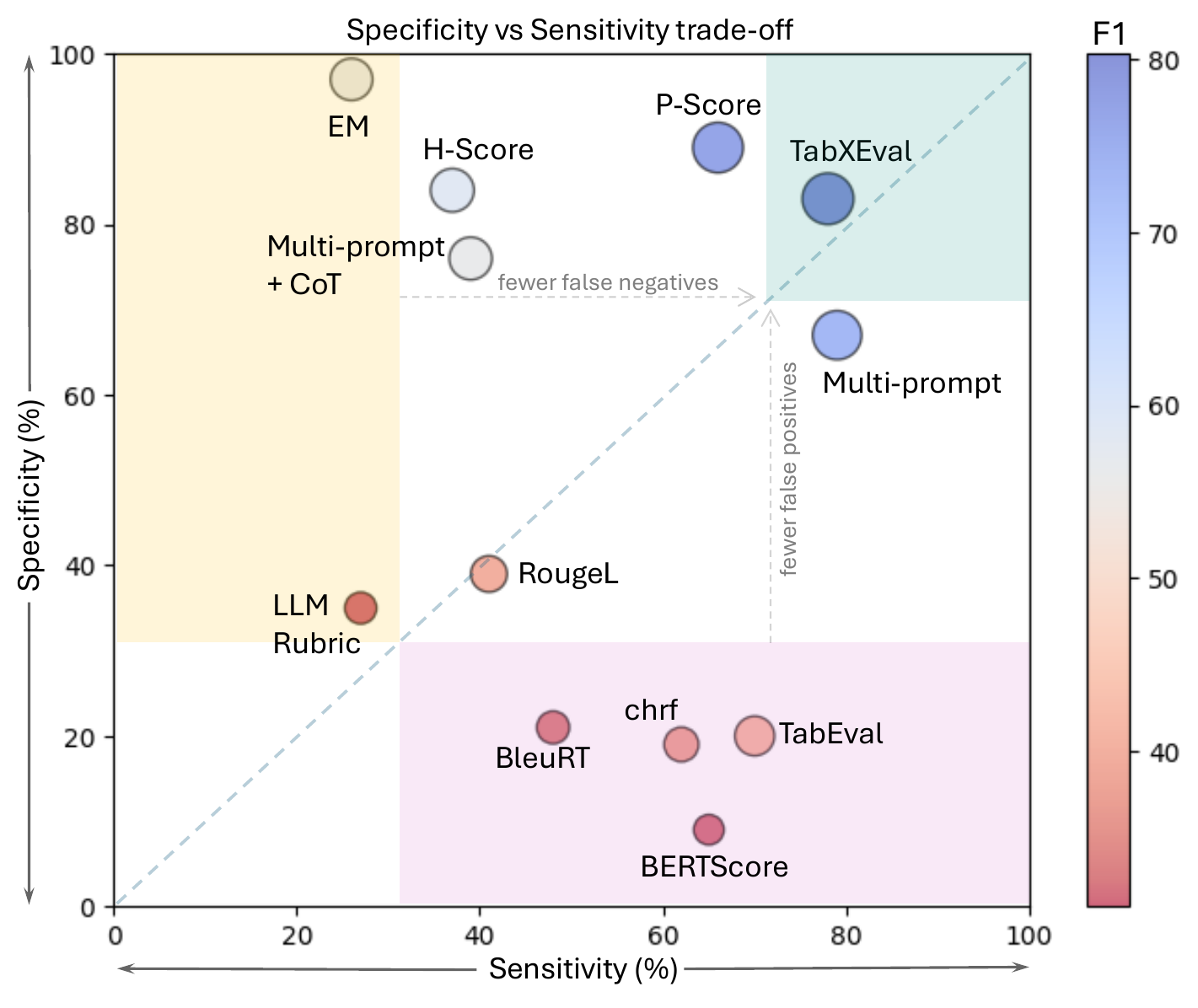} 
    \vspace{-1.5em}
    \caption{Sensitivity–Specificity trade-off across metrics. Bubble size reflects the harmonic mean of accuracy, sensitivity, and specificity; color shows F1-score. The dashed line marks ideal balance top-right methods perform best. Green denotes optimal (\textbf{Goldilocks}) zone; pink favors sensitivity, yellow favors specificity.}
    \label{fig:sensitivity_specificity}
    \vspace{-1.5em}
\end{figure}

\section{Introduction}

Tables are a ubiquitous data format across critical workflows: budget forecasts, patient dashboards, and experimental logs alike where even a one-cell error can trigger costly re-statements or clinical misinterpretations.  
As large language models (LLMs) and other neural systems are increasingly tasked with \emph{generating} or \emph{transforming} such tables, reliable automatic \emph{evaluation} becomes a bottleneck.  

Despite the structured nature of tables, most evaluation metrics treat them as plain text. Metrics like BLEU, ROUGE, METEOR, and chrF rely on n-gram overlap, ignoring row–column alignment and unit consistency~\cite{papineni-etal-2002-bleu,lin-2004-rouge,banerjee-lavie-2005-meteor,popovic-2015-chrf}. Embedding-based scores such as \textsc{BERTScore} improve semantic sensitivity but overlook structural errors like column swaps~\cite{bert_score}. Token-level metrics, including Exact Match and \textsc{PARENT}, address factual grounding but fail under reordered or merged schemas~\cite{parent}. Structural benchmarks highlight these issues: \textsc{StructBench} exposes failures on partial cell mismatches~\cite{gu2024structextevalevaluatinglargelanguage}, \textsc{TanQ} reveals brittleness under unit conversions~\cite{tanq}, and \textsc{Data-QuestEval} sacrifices structure for corpus-level QA-based comparisons~\cite{dataquesteval}. Atomic decomposition methods like “Is this a bad table?” improve detection but add opacity and computational cost~\cite{tabeval}. In contrast, work like \textsc{THumB} shows the benefit of rubric-based human ratings in improving evaluation transparency~\cite{thumb}. 


Taken together, existing metrics tend to emphasize either semantics or structure, but rarely both. They offer limited diagnostic insight, often masking specific error types and failing to provide actionable feedback for model improvement. As shown in image-captioning work like \textsc{THumB}, coupling automatic scores with rubric-based human ratings yields more interpretable and reliable evaluations~\cite{thumb}. These findings underscore the need for a rubric-based evaluation framework that explicitly assesses both structural alignment and semantic fidelity. In contrast to single-score metrics that collapse diverse errors—such as schema mismatches, contextual omissions, or subtle content shifts—into a single value, a rubric-based approach offers fine-grained, interpretable feedback. Such granularity is essential for complex or high-stakes tasks, where even minor discrepancies can significantly affect downstream performance.

To overcome these challenges, we introduce \textbf{\tabx{}}, a novel evaluation framework built on a structured, multi-level rubric that combines high-level structural descriptors with fine-grained contextual signals. \tabx{} operates in two phases: \textit{TabAlign} first performs precise alignment of table elements using both rule-based and LLM-assisted strategies, followed by \textit{TabCompare}, which conducts detailed semantic and syntactic analysis over the aligned cells. This design allows \textbf{\tabx{}} to capture both table-level and cell-level discrepancies that prior metrics often overlook.

To rigorously test our rubric and framework, we construct \textbf{\tabbench{}}, a diverse, synthetic benchmark that emulates realistic table perturbations across multiple domains. \tabbench{} includes human-annotated ratings grounded in our rubric, serving as a gold standard for evaluating metric sensitivity, specificity, and alignment with human judgment. By providing controlled, interpretable scenarios, \tabbench{} fills a critical gap in current evaluation practice, enabling robust and explainable assessment of structured table outputs. Unlike prior metrics, \textbf{\tabx{}}, as shown in Figure~\ref{fig:sensitivity_specificity}, excels at detecting subtle discrepancies i.e. sensitive and accurately localizing errors between tables i.e. specific enough. We summarize our main contributions below:



\begin{itemize}
    \item We introduce the first rubric integrating multi-level structural descriptors and fine-grained contextual quantification for robust table comparisons.
    \item We propose \textbf{\tabx{}}, a two-phase LLM-based table evaluation method that aligns reference tables structurally and compares them semantically and syntactically via our rubric.
    \item We construct \textbf{\tabbench{}}, a diverse benchmark derived from multi-domain datasets, validating evaluation metrics through structured perturbations and human assessments.
    \item We analyze the strengths and weaknesses of existing evaluation methods via Sensitivity-Specificity Trade-off.
    \item We present \tabx{}'s qualitative and quantitative effectiveness in table generation task, enabling explainable automatic evaluation.
\end{itemize}


\section{\tabx{}}
We establish a transparent and systematic protocol for table evaluation by comparing a reference table—the candidate table produced by a human or an LLM with a ground-truth table. These tables may differ in formatting, interpretation, or unit representation, necessitating a rigorous evaluation framework. For instance, an LLM-generated table might omit entire rows or abbreviate numeric values (e.g., “100k” vs. “100,000”), while a human-curated table may specify units only in the header (e.g., “velocity” vs. “velocity (m/s)”). To address such discrepancies, we design a set of rules based rubrics \tabx{}, which aim to improve the reliability and interpretability of table evaluation.

\begin{figure*}[ht]
    \centering
    \includegraphics[width=\linewidth]{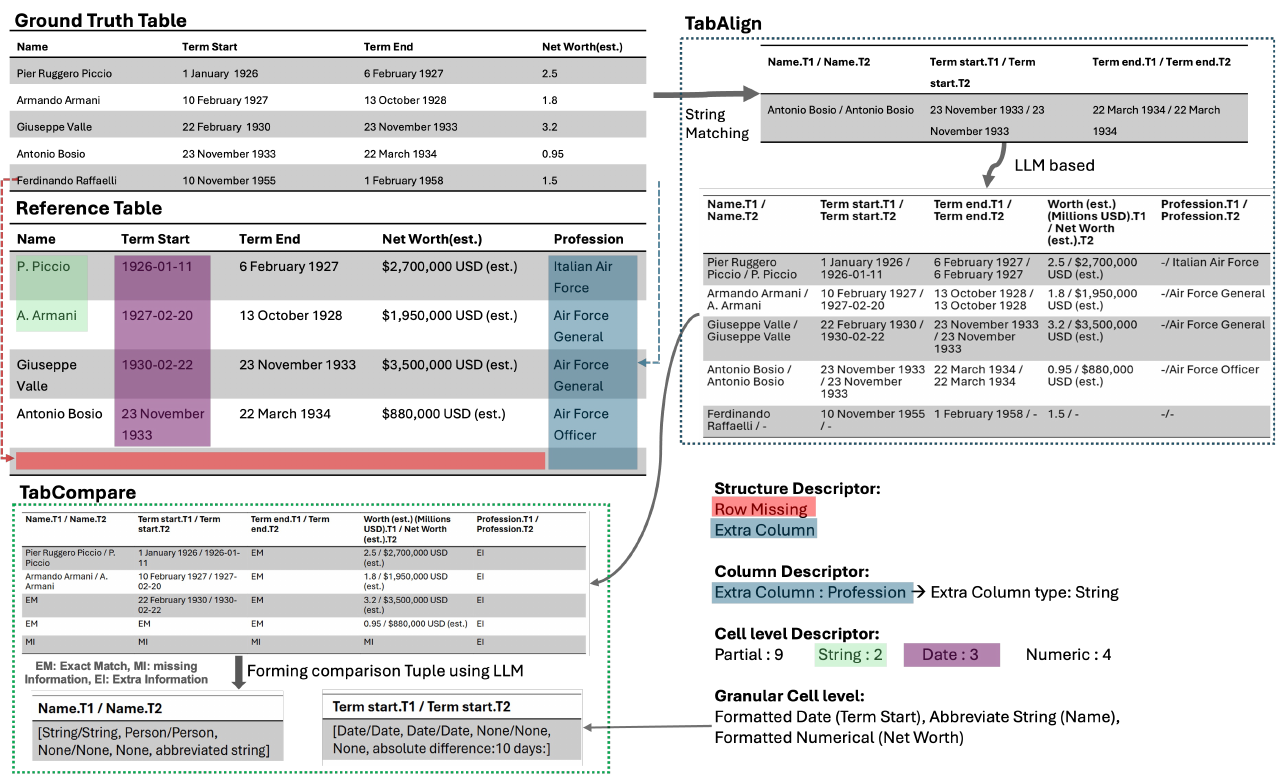} 
    \vspace{-1.5em}
    \caption{
  End-to-end schematic of \textbf{\tabx}. (1) \emph{TabAlign} aligns rows, columns, and cells using deterministic rules plus an LLM refinement loop. (2) \emph{TabCompare} classifies each aligned cell as extra, missing, or partial and combines the counts with rubric weights~$(\alpha,\beta,\gamma)$. This workflow populated the rubrics and outputs a table-level score and cell-level error trace, enabling fine-grained analysis.}
    \label{fig:main}
    \vspace{-1.0em}
\end{figure*}







\subsection{Evaluation Rubric}
\label{sec:rubric}
Towards ensuring consistency and fairness in structure evaluation we provide an exhaustive rubric that enhances clarity and transparency in reference-based table evaluation. To quantify the degree of correctness and coverage of information at the most granular level, we advocate 4 categories of evaluation protocols:

\vspace{0.25em}
\noindent  \textbf{Structure Descriptor} coarsely assesses the overall structure against ground truth. This component compares information at the table level (missing information, extra information and exact matches), giving a high level description of information integrity.

\vspace{0.5em}
\noindent \textbf{Column Descriptor} identifies data types of each column based on missing and extra information. This allows for a fine-grained strategy for evaluating cell values, as tables with heterogeneous columns like dates, numbers, strings etc should not be penalized on the same scale.

\vspace{0.25em}
\noindent \textbf{Cell Level Descriptor} looks at both semantic and syntactical representation of cell values. Other methods such as \cite{bert_score, tabeval, gu2024structextevalevaluatinglargelanguage} fail in recognizing different representations of the same data and hence can wrongly penalize correct information. This component of the rubric is necessary to craft an ideal table evaluation framework.

\vspace{0.25em}
\noindent \textbf{Granular Cell Level  Difference}  that determines the magnitude of discrepancies between reference and ground truth table necessary to quantify instances w.r.t. cell level descriptions.
It also captures variations by changing the format(e.g. m to cm, years to date) to report the absolute differences.

\subsection{\textbf{\tabx{}} Rubric}
\label{sec:method}

We propose \textbf{\tabx{}}, a two-phase framework that combines deterministic rules and LLM-based analysis for robust and interpretable table evaluation. This design strikes a balance between precision (capturing exact matches) and flexibility (handling semantic or structural variations).

\paragraph{Phase 1: \textit{TabAlign}} matches columns, rows, and cells between the reference and candidate tables. We begin with exact string matching to establish a precise baseline alignment. Next, we refine this alignment using an LLM to account for abbreviations, synonyms, and structural transformations (e.g., merged columns, row/column transpositions). Purely exact matching can be overly strict, missing semantically equivalent but syntactically different cells. The LLM-driven refinement ensures a more comprehensive alignment while preserving high precision. Finally, we get an output table which has a combination of strict and relaxed mapping as shown in Figure~\ref{fig:main}.

\paragraph{Phase 2: \textit{TabCompare}} performs a fine-grained evaluation of the aligned tables.
From the refined alignment, we extract table-level statistics (e.g., missing/extra rows or columns) and focus on partially matched cells. These cells are compared in detail using LLM-generated ``comparison tuples'' as shown in Figure~\ref{fig:main} which capture numeric, string, date/time, and unit mismatches. We also compute magnitudes of differences (e.g., converting months to days) for precise reporting of discrepancies. Table-level summaries alone cannot uncover subtle cell-level errors, such as unit mismatches or minor numeric discrepancies. By combining table-level statistics with granular cell comparisons, \tabx{} yields a more reliable and transparent assessment of content fidelity.

\paragraph{Score}

Our scoring function for \tabx{} is defined as follows:

\vspace{-1.5em}
\begin{equation*}
\label{eq:score-main}
\begin{aligned}
\mathbf{\tabx{}}
&= \sum_{I\in\{\text{Missing},\,\text{Extra},\,\text{Partial}\}}
     \beta_{I} \\
&\quad\times
     \Bigl(\sum_{E\in\{\text{row},\,\text{column},\,\text{cell}\}}
            \alpha_{E}\,\tfrac{f_{E}}{N_{E}}
     \Bigr)
     \gamma_{p}\,.
\end{aligned}
\end{equation*}

where $\beta_{I}$ is the weight assigned to each type of information error ($I$) such as Missing, Extra, and Partial; $\alpha_{E}$ is the weight for each entity type ($E$) including rows, columns, and cells; $f_{E}$ represents the number of correctly matched entities; and $N_{E}$ is the total number of entities in the ground truth.

For partial matches at the cell level, the modifier $\gamma_{p}$ is defined as:

\vspace{-1.5em}
\begin{equation*}
\label{eq:score-partial}
\gamma_{p} =
\begin{cases}
1, & \text{if no partial cell},\\[1mm]
\omega_{p}\,\bigl\lvert\tfrac{GT - Ref}{Ref}\bigr\rvert, 
    & \text{if partial cell detected}.
\end{cases}
\end{equation*}

This formulation captures the multi-level nature of table evaluation insipred by proposed rubric. First aggregating errors across different information types (Missing, Extra, Partial) via the outer summation, and then evaluating the correctness at various entity levels (row, column, cell) using the inner summation. The term $\gamma_{p}$ further refines the score by quantifying discrepancies in partially matched cells through a normalized absolute difference between the ground truth ($GT$) and the reference ($Ref$), ensuring that both coarse structural errors and fine-grained content differences are robustly accounted for in a single, interpretable metric. An illustrative example demonstrating the application of the above equations is provided in Appendix~\ref{sec:eq-example}.

Overall, \tabx{}’s two-phase structure ensures that both coarse (table-level) and fine (cell-level) differences are captured, providing an adaptable and explainable approach to table evaluation.

\subsection{\tabbench{} Benchmark}
\label{sec:tabxbench}

Evaluating table metrics across diverse domains and error types remains a significant challenge due to the limited scope of existing datasets, which often focus on a single domain or contain only select data types. To bridge this gap, we introduce \textbf{\tabbench{}}, a controlled multi-domain test bed that comprehensively captures real-world nuances of tabular data generation and evaluation.

\begin{figure}[h]
    \centering
    \includegraphics[width=\linewidth]{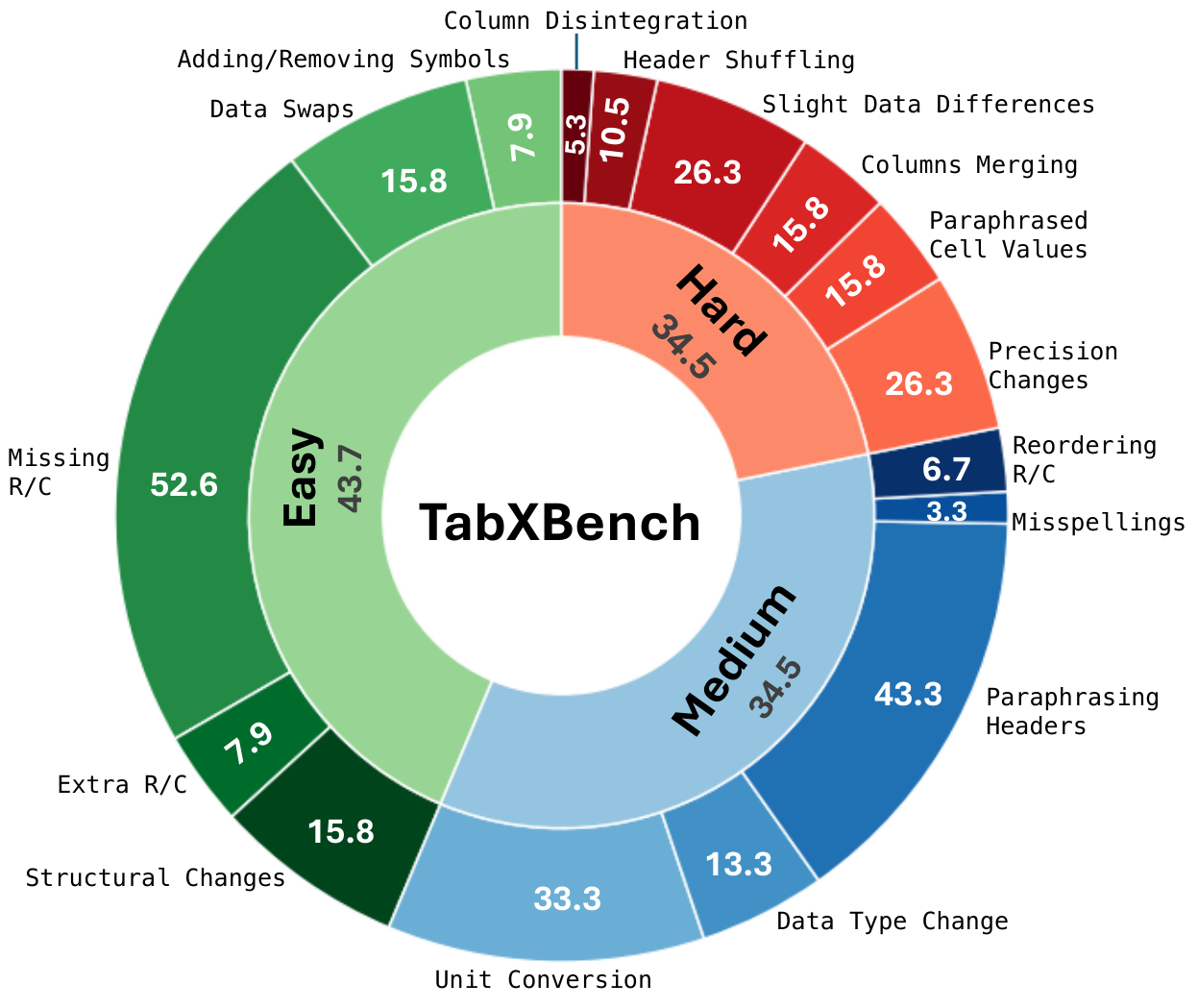}
    \vspace{-1.0em}
 \caption{%
    \textbf{Perturbation spectrum in \tabbench{}.}  
    The outer ring enumerates the frequency (numeric labels) of the 16 fine-grained perturbation types applied to reference tables.  
    The inner ring groups these edits into three difficulty bands \emph{Easy} (light green, $\approx$44\%), \emph{Medium} (blue, $\approx$34\%), and \emph{Hard} (red, $\approx$35\%).  }
    \label{fig:tabxbench}
    \vspace{-1.0em}
\end{figure}

\tabbench{} is designed to rigorously assess the {sensitivity and specificity} of reference-based table evaluation methods. Unlike prior datasets that narrowly target specific tasks (e.g., finance or sports), our benchmark spans multiple domains (finance, sports, knowledge bases, and more) while systematically incorporating an extensive range of potential table perturbations. This diversity ensures that \tabbench{} captures common pitfalls such as missing rows/columns, reordered headers, unit mismatches, numeric discrepancies, and complex structural variations (e.g., row/column transposition).

\begin{table}[!h]
    \centering
    \renewcommand{\arraystretch}{1.1} 
    \setlength{\tabcolsep}{3pt} 
    
    \resizebox{\linewidth}{!}{ 
        \begin{tabular}{l @{\hskip 10pt} c @{\hskip 10pt} c @{\hskip 10pt} c @{\hskip 10pt} c @{\hskip 10pt} c}
            \hline \hline
            \textbf{Dataset} & \textbf{\# of Tables} & \textbf{\# Perturb/Table} & \multicolumn{3}{c}{\textbf{Average}} \\
            \cline{4-6}
             &  &  & \textbf{Rows} & \textbf{Cols} & \textbf{Cells} \\
            \hline
            FINQA      & 13 & 5 & 13.08 & 5.54 & 71.00 \\
            TANQ       & 8  & 5 & 8.38  & 5.38 & 44.25 \\
            ROTOWIRE   & 12 & 5 & 12.08 & 14.08 & 165.67 \\
            FETAQA     & 7  & 5 & 15.86 & 5.14 & 88.14 \\
            WIKITABLES & 5  & 5 & 16.40 & 3.60 & 53.60 \\
            WIKISQL    & 6  & 5 & 8.33  & 5.17 & 42.17 \\
            \hline \hline
        \end{tabular}
    }
    \vspace{-0.5em}
    \caption{Composition of the \tabbench{} corpus across six source datasets.}
    \label{tab:dataset_stats}
    \vspace{-1.0em}
\end{table}

\tabbench{} consists of 50 handpicked ``clean'' (reference) tables from six popular text-to-table and table QA datasets: RotoWire~\cite{rotowire}, TANQ~\cite{tanq}, FetaQA~\cite{nan2021fetaqafreeformtablequestion}, FinQA~\cite{finqa}, WikiTable ~\cite{wikitables}, and WikiSQL~\cite{wikisql} and the statistics can be found in \autoref{tab:dataset_stats}. Each table was augmented with five distinct perturbations spanning over 16 error types, carefully curated to reflect common generation mistakes. The perturbations were first drafted with LLM assistance (e.g., reformatting numeric values, altering units, swapping headers) and subsequently validated by human experts for correctness and variety. We categorize the resulting perturbed tables into three difficulty levels (Easy, Medium, Hard) as demonstrated in \autoref{fig:tabxbench}, ensuring coverage of both straightforward errors (e.g., minor typos) and complex structural manipulations (e.g., merged cells or shifted rows).

By offering a controlled yet diverse environment, \tabbench{} enables:
\textbf{Fine-Grained Analysis}: Researchers can systematically evaluate how well metrics detect specific error types (e.g., unit mismatches vs. missing rows). \textbf{Sensitivity-Specificity Trade-offs}: The benchmark’s difficulty levels and variety of perturbations allow detailed insights into a method’s robustness and tolerance for minor vs. major errors. \textbf{Realistic Scoring Correlations}: It includes human annotations (aligned with our rubric in Section~\ref{sec:rubric}), enabling correlation studies that compare machine-generated scores to human judgments.

Finally, to illustrate \tabx{} pragmatic value, we apply it to evaluate table outputs from three LLMs across four standard tasks - RotoWire, TANQ, WikiBio, and WikiTable highlighting how each evaluation method fares in a realistic setting. As a pioneering multi-domain resource, \tabbench{} thus provides a solid foundation for advancing research in reliable, explainable table evaluation.

\section{Experiments}
To validate efficacy of \tabx{}, we conduct experiments using our synthetic dataset \tabbench{}. We report \texttt{GPT-4o}~\cite{openai2024gpt4ocard} and \texttt{LLaMA-3.3-INSTRUCT} results for our framework \tabx{} for both components \textit{TabAlign} and \textit{TabCompare}. 

\vspace{0.5em}
\noindent \textbf{Baselines.} Our evaluation compares \tabx{} against a broad range of baselines, which we classify into deterministic and non-deterministic approaches. Deterministic metrics (e.g., Exact Match (\textsc{EM}), \textsc{chrf}, \textsc{ROUGE-L}) yield fixed outputs based on string- or character-level comparisons, ensuring reproducibility. In contrast, non-deterministic metrics (e.g., \textsc{BERTScore}, \textsc{BLUERT}, \textsc{H-Score}) leverage contextualized neural representations to capture subtle linguistic nuances albeit with potential variability. We also include two recent methods: \textsc{P-Score}, an LLM-based metric outputting scores on a 0–10 scale, and \textsc{TabEval}, an embedding-based method that unrolls tables using an LLM and computes entailment via RoBERTa-MNLI. For a fair comparison, we further propose a Direct-LLM baseline that is prompted with the same evaluation rubric detailed in Section~\ref{sec:rubric}. 

\vspace{0.5em}
\noindent \textbf{LLMs.} Throughout our experiments, we used \texttt{GPT-4o}, \texttt{Gemini 2.0-flash}, and \texttt{LLaMA-3.3-Instruct 70B}. All models were executed with identical sampling settings (default temperature, top‑$k$, and top‑$p$) unless specified otherwise. All our prompts for TabAlign, TabCompare and Direct-LLM basline are given in Appendix~\ref{sec:prompts}.

\subsection{Human correlation}
For each sample in the ground truth set of \tabbench{}, two human evaluators evaluated the tables using our proposed rubrics and guidelines. For both Human and \tabx{}, we present ground-truth and a randomly selected perturbation (out of 5) to fill the rubrics. The detailed human annotation protocol is described in Appendix~\ref{sec:Human-Eval}. Finally these rubrics are compared using both Pearson~\cite{pearson} and Kendall’s Tau~\cite{kendals} correlation coefficients to quantify the degree of alignment between our method and human ratings. We observe very high correlation, i.e, $\mathbf{99.7\%}$ and $\mathbf{95.1\%}$ Pearson's $\rho$ correlation for the Rubric Structure Descriptor and Cell Level Descriptor respectively. 

In contrast, Direct LLM based baseline correlates, only $\mathbf{30.6\%}$ and $\mathbf{40.6\%}$ Pearson's $\rho$ correlation for the Rubric Structure Descriptor and Cell Level Descriptor respectively. Revealing that it fails to understand the rubric, and quantify the structural and contextual challenges in table evaluation, and hence is unable to correctly align with humans correlation. Similarly we report $\mathbf{99.1\%}$ and $\mathbf{92.8\%}$ human correlation using Kendall's $\tau$. While the baseline is $\mathbf{30.7\%}$ and $\mathbf{55\%}$ on Structure Descriptor and Cell Level Descriptor rubric respectively. These results demonstrate that disentangling the alignment and comparison phases is critical for robust table evaluation as Direct method falls short even hen presented with evaluation rubrics. \tabx{}'s correlation with human are consistent in capturing multi-level nuances for real-world challenges in table evaluation resembling humans.

\subsection{Human Ranking Correlation Study}

To further validate the robustness of our evaluation framework, we conducted a human ranking correlation study using outputs from \tabbench{}. In this study, expert annotators ranked the quality of a ground-truth table and its perturbed variants each reflecting real-world errors such as structural, semantic, and formatting issues. These human rankings serve as our gold standard for assessing table quality.

For each table, we computed an aggregated score (based on cell-level f\(_1\) measures) using various evaluation metrics, including deterministic baselines (e.g., Exact Match, chrf, ROUGE-L) and non-deterministic methods (e.g., BERTScore, H-Score, P-Score, BLUERT, TabEval), alongside our proposed \tabx{}. Further, for baselines implementation we run both a single-step and multi‐step LLM baseline to populate our proposed rubric tables that mirrors our two‐stage \tabx{} pipeline more closely along with LLM based ranking and multi‐step LLM baseline with Chain-of-thoughts.
We then measured the correlation between the automatic rankings and the human judgments using multiple metrics: Spearman’s \(\rho\), Kendall’s \(\tau\), Weighted Kendall’s \(\tau^{\dagger}\), Rank-Biased Overlap (RBO), and Spearman’s Footrule. These measures collectively assess both the overall ranking order and positional differences.

\begin{table*}[t]
\small
\centering
\begin{tabular}{lccccc}
\hline \hline
\multirow{1}{*}{\textbf{Metrics}} &
  \textbf{Spearman's $\rho$ $\uparrow$} &
  \textbf{Kendall's $\tau$ $\uparrow$} &
  \textbf{W-Kendall's $\tau^{\dagger}$ $\uparrow$} &
  \textbf{RBO $\uparrow$} &
  \textbf{Spearman's \textbf{Footrule} $\downarrow$} \\
  \hline
\textbf{\textsc{EM}}         & 0.18  & 0.16  & 0.16  & 0.26 & 0.57 \\
\textbf{\textsc{chrF}}       & 0.12  & 0.11  & 0.08  & 0.25 & 0.59 \\
\textbf{\textsc{H-Score}}    & 0.14  & 0.11  & 0.09  & 0.28 & 0.51 \\
\textbf{\textsc{BERTScore}}  & 0.19  & 0.15  & 0.13  & 0.25 & 0.57 \\
\textbf{\textsc{ROUGE-L}}    & 0.21  & 0.18  & \textbf{0.40}  & 0.27 & 0.53 \\
\textbf{\textsc{BLEURT}}     & 0.29  & 0.25  & 0.25  & 0.27 & 0.51 \\
\textbf{\textsc{TabEval}}    & -0.04 & -0.04 & -0.03 & 0.23 & 0.63 \\
\textbf{\textsc{P-Score}}    & \underline{0.30}  & \underline{0.27}  & 0.24  & \underline{0.31} & \underline{0.39} \\
\textbf{\textsc{LLM rubric}}   & 0.23  & 0.16  & 0.17  & 0.28 & 0.47 \\
\textbf{\textsc{LLM ranking}}   & 0.29  & 0.24  & 0.23  & 0.30 & 0.41 \\
\textbf{\textsc{Multi-prompt }}   & 0.29  & 0.24  & 0.23  & 0.30 & 0.42 \\
\textbf{\textsc{Multi-prompt + CoT}}   & 0.30  & 0.25  & 0.24  & 0.29 & 0.45 \\ \hline
\textbf{\tabx{}}   & \textbf{0.44}  & \textbf{0.40}  & \underline{0.38}  & \textbf{0.34} & \textbf{0.29} \\ \hline \hline
\end{tabular}%
\vspace{-0.75em}
\caption{Correlation between automatic rankings and human judgments. Higher Spearman’s \(\rho\), Kendall’s \(\tau\), Weighted Kendall’s \(\tau^{\dagger}\), and RBO values indicate better agreement, while lower Spearman’s Footrule values are preferable.}
\vspace{-1.75em}
\label{tab:human}
\end{table*}

As shown in \autoref{tab:human}, \tabx{} achieves the strongest correlation with human rankings across all metrics. Specifically,

\vspace{0.5em}
\noindent \textbf{Overall Ranking Order:} \tabx{} attains a Spearman’s \(\rho\) of 0.44 a relative improvement of nearly 47\% over the next-best method (P-Score at 0.30). Its Kendall’s \(\tau\) of 0.40 and Weighted Kendall’s \(\tau^{\dagger}\) of 0.38 further indicate strong monotonic agreement with human assessments.

\vspace{0.5em}
\noindent \textbf{Top-Weighted Agreement:} With an RBO of 0.34, \tabx{} demonstrates superior alignment in the higher-ranked items, compared to values ranging from 0.23 to 0.31 for other methods.

\vspace{0.5em}
\noindent \textbf{Positional Accuracy:} \tabx{} records the lowest Spearman’s Footrule distance (0.29), reflecting minimal positional discrepancy relative to the human gold standard.

Notably, the TabEval method not only fails to capture these nuances as it negatively correlates, highlighting its inability to account for the multifaceted nature of table quality. This analysis reinforces the importance of our two-phase approach, disentangling structural alignment (TabAlign) from detailed cell-level comparison (TabCompare) to effectively mirror human judgment. Finally, the human ranking correlation study clearly demonstrates that \tabx{} provides a more robust, interpretable, and human-aligned evaluation of table outputs, capturing both coarse and fine-grained discrepancies that are critical in real-world scenarios.

\subsection{What Sets \tabx{} Apart?}

A key criteria of any evaluation metric is to achieve a balance between \textit{specificity} (i.e., avoiding false positives) and \textit{sensitivity} (i.e., avoiding false negatives). In \autoref{fig:sensitivity_specificity}, we visualize this trade-off by plotting each metric’s specificity (y-axis) against its sensitivity (x-axis). The background colormap in the figure corresponds to the F1 score. Finally, Bubble Size represented by harmonic mean of specificity, sensitivity and accuracy, providing a quick visual cue for overall performance. 

\vspace{0.5em}
\noindent \textbf{Goldilocks Zone for Ideal Metrics.} Metrics positioned in the top-right portion of the chart (the green-shaded ``Goldilocks zone’’) demonstrate the desired trait of consistently identifying correct table content (\textit{high sensitivity}) while minimizing the likelihood of falsely flagging errors (\textit{high specificity}). \tabx{} resides firmly in this zone, illustrating its balanced performance across diverse table perturbations.\\

\noindent \textbf{Comparisons with Other Metrics.}
We compare \textbf{\tabx{}} against several widely used metrics:
(1) \textsc{P-Score} performs well at a high level and sits near the Goldilocks zone in our evaluations, reflecting strong table-level correctness. However, it lacks \textit{explainability} and \textit{granular insights}, offering only a single $0$–$10$ score that limits interpretability and error traceability.
(2) \textsc{H-Score} and \textsc{BERTScore} better capture semantics than string-based metrics such as \textsc{EM}, \textsc{ROUGE-L}, and \textsc{chrF}, but often overlook structural errors like swapped columns or missing rows—resulting in moderate sensitivity but poor specificity.
(3) \textsc{TabEval} uses entailment over LLM-generated atomic statements, but frequently misses fine-grained numeric or unit mismatches. In our experiments, it produced false positives on tables that were re-formatted yet semantically equivalent.

\vspace{0.5em}
\noindent \textbf{Why \tabbench{} Matters.}  
\tabbench{} introduces diverse table perturbations (e.g., missing rows/columns, numeric/unit mismatches) to stress-test metrics across real-world errors (\autoref{fig:tabxbench}). Simple metrics like \textit{Exact Match} fail under reordering or semantic shifts, while LLM/embedding-based methods miss unit mismatches or partial errors. \tabx{}’s \textit{two-phase} approach first aligning structure (\textsc{TabAlign}), then systematically comparing content (\textsc{TabCompare}) ensures precise discrepancy detection aligned with human judgment.  

\vspace{0.5em}
\noindent \textbf{Significance of \tabx{}.}  
By balancing sensitivity, specificity, and F1 scores, \tabx{} not only outperforms across evaluation dimensions but also \textit{explains} its judgments via structured rubrics. This transparency is crucial for financial reporting, scientific validation, and knowledge curation, where subtle errors can be costly. Identifying \textit{what} went wrong and \textit{where}, \tabx{} provides both quantitative and qualitative insights for improving table generation.  

In summary, \tabx{}’s interpretability, robust performance, and alignment with \tabbench{}’s challenging setup establish it as the new standard for explainable, human-aligned table evaluation.

\begin{table*}[ht]
\centering
\resizebox{\linewidth}{!}{%
\begin{tabular}{l*{21}{c}}
\hline\hline
 & \multicolumn{7}{c}{\textbf{LLaMA-3.3 70B}} & \multicolumn{7}{c}{\textbf{GPT-4o}} & \multicolumn{7}{c}{\textbf{Gemini-2.0-flash}} \\ \hline
\textbf{Stat} & Num & String & Bool & Date & List & Time & Others & Num & String & Bool & Date & List & Time & Others & Num & String & Bool & Date & List & Time & Others \\ \hline
\multicolumn{22}{c}{\cellcolor[HTML]{ECF4FF}\textit{\textbf{WikiTables}}} \\ \hline
EI      & 0.05 & 4.33 & 0.00 & 0.17 & 0.00 & 0.00 & 0.13  & 0.03 & \cellcolor[HTML]{F88379}\textbf{1.17} & 0.00 & 0.02 & 0.00 & 0.00 & 0.13  & 0.02 & 2.33 & 0.00 & 0.11 & 0.00 & 0.00 & 0.07 \\
MI      & 0.01 & 0.80 & 0.00 & 0.03 & 0.00 & 0.00 & 0.00  & 0.01 & 0.93 & 0.00 & 0.00 & 0.00 & 0.00 & 0.01  & 0.00 & 0.73 & 0.00 & 0.00 & 0.00 & 0.00 & 0.00 \\
Partial & 0.22 & 25.00 & 0.00 & 0.35 & 0.00 & 0.01 & 0.09  & \cellcolor[HTML]{F88379}\textbf{0.32} & 20.34 & 0.00 & 0.55 & 0.00 & 0.02 & 0.10  & 0.30 & 22.50 & 0.00 & 0.48 & 0.00 & 0.02 & 0.07 \\ \hline
\multicolumn{22}{c}{\cellcolor[HTML]{DAE8FC}\textit{\textbf{WikiBio}}} \\ \hline
EI      & 0.04 & 2.84 & 0.00 & 0.09 & 0.00 & 0.00 & 0.12  & 0.04 & 2.02 & 0.00 & 0.03 & 0.00 & 0.00 & 0.09  & 0.04 & 2.30 & 0.00 & 0.07 & 0.00 & 0.00 & 0.06 \\
MI      & 0.02 & 0.29 & 0.00 & 0.00 & 0.00 & 0.00 & 0.00  & 0.00 & 0.38 & 0.00 & 0.01 & 0.00 & 0.00 & 0.00  & 0.00 & 0.37 & 0.00 & 0.01 & 0.00 & 0.00 & 0.00 \\
Partial & 0.16 & 14.38 & 0.00 & 2.60 & 0.00 & 0.00 & 0.03  & 0.16 & 15.80 & 0.00 & 0.86 & 0.00 & 0.00 & 0.03  & 0.15 & 13.59 & 0.00 & 2.97 & 0.00 & 0.00 & 0.04 \\ \hline
\multicolumn{22}{c}{\cellcolor[HTML]{D4FFD4}\textit{\textbf{TANQ}}} \\ \hline
EI      & 0.05 & 0.84 & 0.00 & 0.16 & 0.09 & 0.00 & 0.00  & 0.02 & 0.18 & 0.00 & 0.06 & 0.03 & 0.01 & 0.00  & 0.00 & 0.29 & 0.00 & 0.07 & 0.02 & 0.00 & 0.00 \\
MI      & 0.01 & 0.24 & 0.00 & 0.11 & 0.01 & 0.04 & 0.00  & 0.01 & 0.08 & 0.00 & 0.05 & 0.02 & 0.01 & 0.00  & 0.02 & 0.21 & 0.00 & 0.05 & 0.00 & 0.02 & 0.00 \\
Partial & 2.73 & 20.50 & 0.00 & 4.72 & 4.82 & 2.19 & 0.07  & \cellcolor[HTML]{F88379}\textbf{1.28} & \cellcolor[HTML]{F88379}\textbf{11.92} & 0.00 & 3.48 & 3.46 & 1.32 & 0.01  & \cellcolor[HTML]{F88379}\textbf{1.22} & \cellcolor[HTML]{F88379}\textbf{9.35}  & 0.00 & 2.02 & 2.64 & 1.12 & 0.02 \\ \hline
\multicolumn{22}{c}{\cellcolor[HTML]{FFFFC7}\textit{\textbf{RotoWire}}} \\ \hline
EI      & 0.84 & 0.50 & 0.00 & 0.00 & 0.00 & 0.00 & 0.01  & 0.68 & 0.23 & 0.00 & 0.00 & 0.00 & 0.00 & 0.00  & 1.31 & 0.54 & 0.00 & 0.00 & 0.00 & 0.00 & 0.09 \\
MI      & 0.97 & 0.32 & 0.00 & 0.00 & 0.00 & 0.00 & 0.04  & 0.56 & 0.08 & 0.00 & 0.00 & 0.00 & 0.00 & 0.00  & 1.06 & 0.28 & 0.00 & 0.00 & 0.00 & 0.00 & 0.02 \\
Partial & \cellcolor[HTML]{F88379}\textbf{0.66} & \cellcolor[HTML]{F88379}\textbf{0.92} & 0.00 & 0.00 & 0.00 & 0.00 & 0.00  & \cellcolor[HTML]{F88379}\textbf{0.31} & \cellcolor[HTML]{F88379}\textbf{0.69} & 0.00 & 0.00 & 0.00 & 0.00 & 0.00  & \cellcolor[HTML]{F88379}\textbf{2.72} & \cellcolor[HTML]{F88379}\textbf{3.87} & 0.00 & 0.00 & 0.00 & 0.00 & 0.03 \\ \hline \hline
\end{tabular}%
}
\vspace{-0.5em}
\caption{%
\textbf{Cell-Level Performance Analysis} of Extra (EI), Missing (MI), and Partial mismatches across data types   numerical, string, boolean, date, list, time, and other   for WikiTables, WikiBio, TANQ, and RotoWire. \textbf{Highlights:} GPT-4o shows fewer string EI in WikiTables and lower partial errors in numerical and string cells in TANQ and RotoWire.}\label{tab:cell-stats}
\vspace{-1.0em}
\end{table*}

\begin{table}[ht]
\small
\centering
\resizebox{\linewidth}{!}{%
\begin{tabular}{lccccccccc}
\hline \hline
 &  \multicolumn{3}{c}{\textbf{LLaMA-3.3 70B}} &  \multicolumn{3}{c}{\textbf{GPT-4o}} &  \multicolumn{3}{c}{\textbf{Gemini-2.0-flash}} \\ \hline
 &  \textbf{MI} &  \textbf{EI} &  \textbf{EM} &  \textbf{MI} &  \textbf{EI} &  \textbf{EM} &  \textbf{MI} &  \textbf{EI} &  \textbf{EM} \\ \hline
\multicolumn{10}{c}{\cellcolor[HTML]{ECF4FF}\textit{\textbf{WikiTable}}}             \\ \hline
Row & 8.69  & 15.82 & 25.59 & 23.44 & 11.51 & \cellcolor[HTML]{F88379}\textbf{27.11} & 20.81 & 10.67 & 26.03 \\
Col & 0.37  & 0.92  & 1.67  & \cellcolor[HTML]{F88379}\textbf{4.47}  & \cellcolor[HTML]{F88379}\textbf{0.07}  & 1.02  & 2.97  & 0.37  & 1.55  \\ \hline
\multicolumn{10}{c}{\cellcolor[HTML]{DAE8FC}\textit{\textbf{WikiBio}}}               \\ \hline
Row & 25.16 & 29.33 & 16.17 & 26.09 & 27.63 & 19.39 & 30.08 & 24.38 & 16.89 \\
Col & 0.10  & 0.0   & 0.05  & 0.05  & 0.025 & 0.0   & 0.12  & 0.0   & 0.0   \\ \hline
\multicolumn{10}{c}{\cellcolor[HTML]{D4FFD4}\textit{\textbf{TANQ}}}                  \\ \hline
Row & 7.6   & 5.83  & 10.97 & 8.27  & 2.80  & 13.00 & 8.51  & 4.01  & 13.01 \\
Col & 2.69  & 1.82  & 23.89 & 2.24  & 0.19  & 22.42 & 2.82  & 0.63  & 21.78 \\ \hline
\multicolumn{10}{c}{\cellcolor[HTML]{FFFFC7}\textit{\textbf{RotoWire}}}              \\ \hline
Row & 3.48  & 39.22 & 17.62 & 1.32  & 28.65 & 38.41 & 3.10  & 22.82 & 13.80 \\
Col & 10.87 & 16.21 & 52.70 & 17.57 & 5.71  & 60.24 & 16.35 & 10.52 & 48.51 \\ 
\hline \hline
\end{tabular}%
}
\vspace{-0.5em}
\caption{\textbf{Table-Level Performance Analysis}: Row/Column MI, EI, and EM rates on WikiTables, WikiBio, TANQ, and RotoWire. \textbf{Highlights:} GPT-4o leads with highest \underline{Row EM (27.11)} and lowest \underline{Col EI (0.07)} on WikiTables.}
\vspace{-1.5em}
\label{tab:table-stats}
\end{table}

\subsection{Performance Analysis}

The results from our Endurance Test on Text-to-Table Generation as depicted in \autoref{tab:cell-stats} and \autoref{tab:table-stats} clearly demonstrate how \tabx{}'s two-phase evaluation framework enables us to drill down from overall table structure to fine-grained cell details.

\vspace{0.5em}
\noindent \textbf{Table-Level Performance}
Across datasets, \texttt{GPT-4o} frequently exhibits higher exact match scores (EM) for both rows and columns compared to \texttt{LLaMA-3.3} and \texttt{Gemini-2.0-flash}. For instance, on WikiTables, \texttt{GPT-4o} achieves a row EM score of 27.11, surpassing the performance of the other models. Additionally, \texttt{GPT-4o} consistently maintains low Extra Information (EI) values at the column level (e.g., only 0.07 EI on WikiTables), indicating that it preserves the intended table structure with minimal unintended additions. Such metrics underscore the model's ability to capture overall table integrity across diverse datasets, including WikiBio, TANQ, and RotoWire.

\vspace{0.5em}
\noindent \textbf{Cell-Level Performance}
A closer examination at the cell level reveals further nuances. In datasets such as WikiTables and WikiBio, GPT-4o records significantly fewer errors in string cells; for example, its string EI on WikiTables is only 1.17 compared to 4.33 for \texttt{LLaMA-3.3}. On TANQ, both \texttt{GPT-4o} and \texttt{Gemini-2.0-flash} show lower partial errors in numerical and string cell types relative to \texttt{LLaMA-3.3}, suggesting more robust semantic and syntactic matching. Notably, on RotoWire, \texttt{GPT-4o} also demonstrates lower partial error counts in both numerical and string cells when compared with \texttt{Gemini-2.0-flash}. These detailed cell-level insights are crucial as they highlight the models' abilities to handle fine-grained discrepancies such as unit mismatches or subtle formatting errors. This course-to-fine grain evaluation not only facilitates the identification of specific error types but also offers interpretable insights into the strengths and weaknesses of each model.

\begin{table*}[h]
\small
\centering
\begin{tabular}{lccccc}
\hline \hline
\multirow{1}{*}{\textbf{Metrics}} &
  \textbf{Spearman's $\rho$ $\uparrow$} &
  \textbf{Kendall's $\tau$ $\uparrow$} &
  \textbf{W-Kendall's $\tau^{\dagger}$ $\uparrow$} &
  \textbf{RBO $\uparrow$} &
  \textbf{Spearman's \textbf{Footrule} $\downarrow$} 
  \\
  
\hline
P-Score (GPT-4o)               & 0.30 & 0.27 & 0.24 & 0.31 & 0.39 \\
BleuRT                         & 0.29 & 0.25 & 0.25 & 0.27 & 0.51 \\
\tabx{} (GPT-4o)               & \textbf{0.44} & \textbf{0.40} & \textbf{0.38} & \textbf{0.34} & \textbf{0.29} \\
\tabx{} (LLaMA)                & 0.37 & 0.30 & 0.30 & 0.29 & 0.44 \\
\tabx{} (Qwen)                 & 0.33 & 0.27 & 0.28 & 0.30 & 0.38 \\
\tabx{} (Gemini)               & 0.30 & 0.23 & 0.21 & 0.30 & 0.40 \\
\hline \hline
\end{tabular}%
\vspace{-0.5em}
\caption{%
    \textbf{Robustness of \tabx{} across LLM back-bones.}  
    Human-ranking correlations for \tabx{} run with \texttt{GPT-4o}, \texttt{LLaMA-3.3 70B-Instruct}, \texttt{Qwen-72B-Instruct} and \texttt{Gemini-2.0-pro}, compared to BLEURT and the P-Score baseline.  \tabx{} + \texttt{GPT-4o} attains the strongest alignment (Spearman’s $\rho$ = 0.44), but all back-bones retain a $\ge$ 0.30 correlation, indicating backbone-agnostic reliability.}
    \vspace{-1.0em}
\label{tab:ranking_metrics}
\end{table*}

\begin{figure}[t]
    \centering
    \includegraphics[width=1\linewidth]{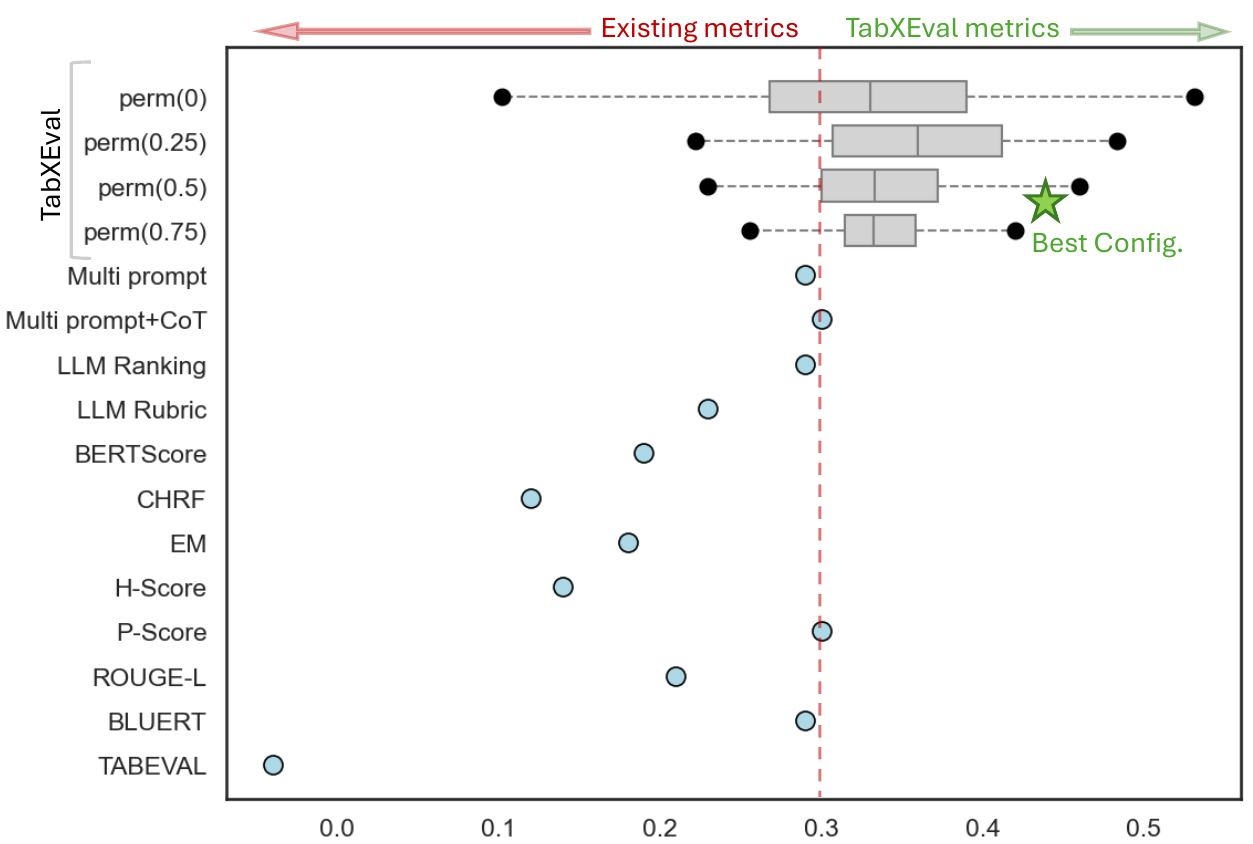} 
     \vspace{-0.75em}
     \vspace{-0.5em}
    \caption{Human Ranking Correlation. This plot compares existing metrics (with the best performance indicated by the red dashed line) against various configurations of our \tabx{} metric. The green star highlights the best performing \tabx{} configuration.}
     \vspace{-1.5em}
    \label{fig:adaptability_robustness}
\end{figure}


\vspace{0.5em}
\noindent \textbf{Adaptability and Robustness Evaluation}  
\autoref{fig:adaptability_robustness} illustrates that \tabx{} consistently outperforms existing metrics in human-correlation across a broad range of weighting schemes (Appendix \ref{sec:performan_analysis} Figure \ref{fig:tabxbench_perm}). Each box plot corresponds to 2\(^6\) permutations of weights for key dimensions (e.g., missing, extra, row, column, cell, partial); these weights contribute to Equations~\ref{eq:score-main} and~\ref{eq:score-partial}, set either as perm(0)=\(\{0,1\}\) or perm(0.25)=\(\{0.25,1\}\). This flexibility allows \tabx{} to adapt to dataset-specific priorities (e.g., penalizing missing rows more heavily) while maintaining robust, domain-agnostic performance. Notably, our best-performing configuration (marked by the star) demonstrates both high average correlation and low variance, underscoring \tabx{}’s capacity to balance fine-grained descriptors with overall structural fidelity.

Complementing this, \autoref{tab:ranking_metrics} demonstrates \tabx{}’s backbone-agnostic robustness: it achieves strong alignment with human rankings across diverse LLMs, including \texttt{GPT-4o}, \texttt{LLaMA}, \texttt{Qwen}, and \texttt{Gemini}. Even under varying model architectures, \tabx{} maintains a correlation of $\ge$ 0.30 across all metrics, reinforcing its reliability beyond just weighting flexibility. A detailed qualitative analysis, provided in Appendix~\ref{sec:QE}, further demonstrates \tabx{}’s effectiveness across both domain-agnostic and specific weighting configurations.

\section{Comparison with Related Work}
\label{sec:related}

\paragraph{Text-to-Table Generation.} Early text–to–table research exploited single–domain corpora such as \textsc{RotoWire} for basketball summaries~\cite{rotowire}, the \textsc{E2E} restaurant set~\cite{novikova-etal-2017-e2e}, \textsc{WikiBio} infobox–biography pairs~\cite{lebret2016neuraltextgenerationstructured}, and \textsc{WikiTableText}~\cite{wikitables}.  
While pioneering, these resources offer limited schema variety and often encourage hallucinated or under-structured outputs.  
Recent collections address these gaps: \textsc{StructBench} permutes headers, merges columns, and shuffles schemas to test structural generalisation~\cite{gu2024structextevalevaluatinglargelanguage}, whereas \textsc{TanQ} requires multi-hop, multi-source reasoning to generate answer tables~\cite{tanq}.  
Such challenging benchmarks expose systematic weaknesses in both generation models and legacy evaluation metrics, motivating the fine-grained rubric employed by \textsc{TabX}.

\paragraph{Other Evaluation Metrics} \textit{1. Surface and embedding overlap.}
Classic n-gram scores \mbox{BLEU}~\cite{papineni-etal-2002-bleu}, \mbox{ROUGE-L}~\cite{lin-2004-rouge}, \mbox{METEOR}~\cite{banerjee-lavie-2005-meteor}, and \mbox{chrF}~\cite{popovic-2015-chrf} treat a table as flat text, ignoring header alignment or cell hierarchy.  
Embedding-based \textsc{BERTScore}~\cite{bert_score} improves semantic sensitivity, yet still overlooks structural fidelity.  
Token-level Exact~Match and \textsc{PARENT}~\cite{parent} partially reward factual grounding, but cannot detect column swaps or unit shifts.

\textit{2. Structure-aware and reference-less scores.}
\textsc{StructBench} introduces \textsc{H-Score} and \textsc{P-Score}, targeting hierarchical integrity and holistic quality, respectively~\cite{gu2024structextevalevaluatinglargelanguage}.  
\textsc{TabEval} (“Is this a bad table?”) decomposes each table into atomic statements and uses textual entailment to capture fine-grained errors~\cite{tabeval}.  
Complementarily, \textsc{Data-QuestEval} dispenses with references altogether by generating and answering questions directly over the source data, achieving strong human correlation in data-to-text tasks~\cite{dataquesteval}. We further show that our approach performs robustly across a wide range of table structures and can be extended to handle hierarchical formats, as discussed in Section~\ref{sec:hierarchial_table}.

Despite these advances, existing metrics still emphases either semantics or structure and provide a single numeric values with limited error traceability or explainability.  Our two-phase \textsc{TabX} closes this gap by disentangling alignment (\emph{TabAlign}) from cell-level comparison (\emph{TabCompare}), producing an interpretable score that balances sensitivity and specificity.

\section{Conclusion and Future Work}

In this work, we have introduced \tabx{} an eXhaustive and eXplainable, two-phase framework that transforms table evaluation by disentangling structural alignment from detailed cell-level comparison. Our method leverages a comprehensive rubric to quantify both coarse and fine-grained errors, yielding results that strongly correlate with human assessments. By developing \tabbench{}, a challenging multi-domain benchmark with diverse perturbations, we have demonstrated the robustness, explainability, and human-alignment of our approach. While limitations such as computational overhead and handling of hierarchical tables remain, the promising performance of \tabx{} opens avenues for further research and refinement in automatic table evaluation.

We found two key directions for future work. Firstly, while our current approach leverages large language models to ensure robustness and generalization to unseen table structures, their computational overhead can hinder practical deployment. Developing more compact alternatives using smaller models (e.g., BART, T5) would require large-scale, heterogeneous fine-tuning data to preserve performance on out-of-distribution tables—a resource that is currently limited. Future efforts could explore model distillation or semi-supervised training to create lightweight yet reliable variants of \textsc{TabXEval}. Secondly, although \textsc{TabXEval} effectively handles structural variations—such as merged or decomposed cells—through LLM-guided alignment, directly supporting complex hierarchical tables (e.g., nested cells or multi-level headers) remains a challenge. A promising direction is to incorporate a preprocessing step that flattens hierarchical structures into standardized key paths (e.g., transforming “Release Date” with sub-headers “Month” and “Year” into “Release Date.Month” and “Release Date.Year”), enabling compatibility with our approach while preserving semantic structure.

\section*{Limitations}
While \tabx{} demonstrates strong performance, it is not without its limitations. Its reliance on large-scale language models comes at the cost of increased computational overhead, which may impact scalability. Moreover, the method faces challenges when dealing with hierarchical tables, where nested headers and multi-level groupings make alignment significantly more complex. Lastly, as a reference-based evaluation approach, \tabx{} necessitates access to ground-truth tables, leaving the question of referenceless evaluation an open and compelling challenge for future research.

\section*{Ethics Statement}
The authors affirm that this work adheres to the highest ethical standards in research and publication. Ethical considerations have been meticulously addressed to ensure responsible conduct and the fair application of computational linguistics methodologies. Our findings are aligned with experimental data, and while some degree of stochasticity is inherent in black-box Large Language Models (LLMs), we mitigate this variability by maintaining fixed parameters such as temperature, $top_p$, and $top_k$. Furthermore, our use of LLMs, including \texttt{GPT-4o}, \texttt{Gemini}, and \texttt{LLaMA}, complies with their respective usage policies. To refine the clarity and grammatical accuracy of the text, AI based tools such as Grammarly and ChatGPT were employed. Additionally, human annotators who are also among the authors actively contributed to data labeling and verification, ensuring high-quality annotations. To the best of our knowledge, this study introduces no additional ethical risks.

\section*{Acknowledgements}
We thank the Complex Data Analysis and Reasoning Lab at Arizona State University for computational support, and Krishna Singh Rajput for his support during the submission process. Lastly, we thank our lab cat, Coco, for her unwavering commitment to keeping our professor creatively unhinged during deadline season.

\bibliography{custom}

\clearpage
\appendix
\onecolumn

\section*{Appendix}
\section{Performance Analysis}
\label{sec:performan_analysis}

\begin{figure*}[!htb]
    \centering
    \includegraphics[width=1.0\textwidth]{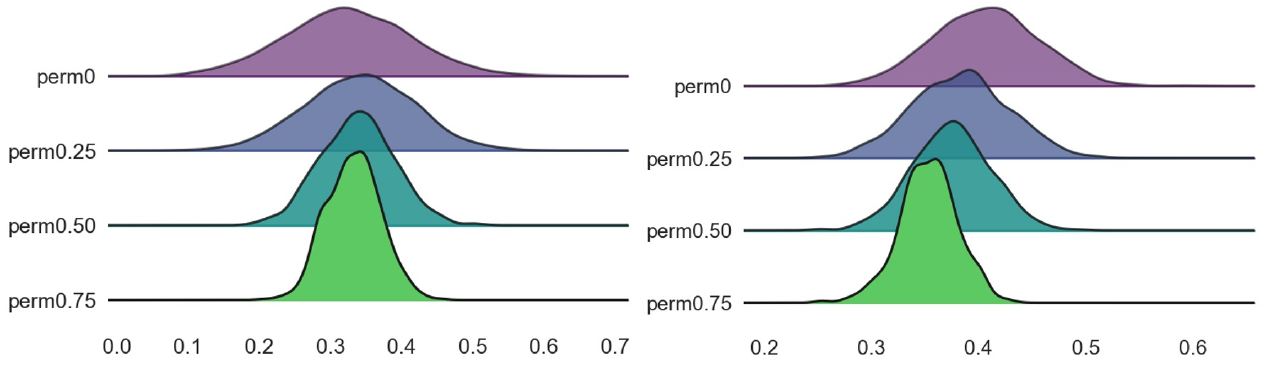}
    \caption{Human Ranking Correlation, ours all configurations.}
    \label{fig:tabxbench_perm}
\end{figure*}

Figure \ref{fig:tabxbench_perm} represent human ranking correlation of \tabx{} across various confirgrations of parameters.

\section{Illustration Example}
\label{sec:eq-example}
Below is a detailed breakdown of our scoring method. For example we are evaluating two movie examples where the Ground Truth Table (GT): 5x5 table and a Reference Table (Ref): 4x6 table.

\begin{equation}
\mathrm{\textbf{\tabx{}}}
= \sum_{I \,\in\,\{\mathrm{Missing},\,\mathrm{Extra},\,\mathrm{Partial}\}}
  \beta_I
  \Biggl(
    \sum_{E \,\in\,\{\mathrm{row},\,\mathrm{column},\,\mathrm{cell}\}}
      \alpha_E \,\frac{f_E}{N_E}
  \Biggr)
  \times \gamma_p
\end{equation}

\section*{Partial Score Weight Definition}

\[
\gamma_p =
\begin{cases}
1, & \text{if no partial cell,}\\[6pt]
\omega_p \times
\displaystyle\left|\frac{GT - Ref}{Ref}\right|,
& \text{if partial detected.}
\end{cases}
\]

\section*{Observed Errors}
\begin{itemize}
  \item \textbf{Missing Row:} The reference table is missing one row compared to the ground truth.
  \item \textbf{Extra Column:} The reference table has an additional column about directors.
  \item \textbf{Partial Match in a Cell:} One cell in the reference table (release date) has a typo and increases the date by 2 days.
\end{itemize}

\section*{Weights}

\subsection*{Entity Weights}
\begin{itemize}
  \item Column: \(\alpha_{\mathrm{column}} = 1\)
  \item Row:    \(\alpha_{\mathrm{row}}    = 0.9\)
  \item Cell:   \(\alpha_{\mathrm{cell}}   = 0.8\)
\end{itemize}

\subsection*{Information Error Weights}
\begin{itemize}
  \item Missing: \(\beta_{\mathrm{Missing}} = 1\)
  \item Extra:   \(\beta_{\mathrm{Extra}}   = 0.9\)
  \item Partial: \(\beta_{\mathrm{Partial}} = 0.8\)
\end{itemize}

\section*{Gamma Modifier for Partial Matches}

For numerical data, we use a weight \(\omega_p = 0.9\) and compute a normalized difference:
\[
\left|\frac{GT - Ref}{Ref}\right| \;=\; 0.4.
\]
Thus,
\[
\gamma_p = 0.9 \times 0.4 = 0.36.
\]
Our proposed rubric is domain agnostic, the beauty of our work is that our weighting scheme is flexible which makes the evaluation metric domain specific based on domain knowledge, for instance one can set higher weights to cell values with numeric types for financial data, in contrast the same rubric can be tuned for sports data where structural nuances should be penalized. Moreover, evidenced by our findings (Figure 4), highlights the strength and generalizability of the underlying scoring mechanism.

\section*{Scoring:}

\subsection*{Missing Row:}
\begin{itemize}
  \item Weight for rows: $\alpha_E = 0.9$
  \item Total rows in GT: 5
  \item Weight for missing: $\beta_I = 1$
\end{itemize}
\[
\text{Contribution to error}
= \alpha_E \times \beta_I \times \frac{\text{Number of missing rows}}{\text{Total rows}}
= 0.9 \times 1 \times \frac{1}{5}
= 0.18
\]

\subsection*{Extra Column:}
\begin{itemize}
  \item Weight for columns: $\alpha_E = 1$
  \item Weight for extra: $\beta_I = 0.9$
  \item Total columns in GT: 5
\end{itemize}
\[
\text{Contribution to error}
= \alpha_E \times \beta_I \times \frac{1}{5}
= 1 \times 0.9 \times \frac{1}{5}
= 0.18
\]

\subsection*{Partial Match (Cell):}
\begin{itemize}
  \item Weight for cells: $\alpha_E = 0.8$
  \item Weight for partial: $\beta_I = 0.8$
  \item Total cells in GT: $5\times5 = 25$
  \item $\gamma_p = 0.36$
\end{itemize}
\[
\text{Contribution to error}
= \alpha_E \times \beta_I \times \frac{1}{25} \times \gamma_p
= 0.8 \times 0.8 \times \frac{1}{25} \times 0.36
= 0.0088
\]

\subsection*{Total Error Score:}
\[
0.18 + 0.18 + 0.0088 = \boxed{0.368}
\]
By adjusting the weights assigned to entity types (rows, columns, cells), information errors (missing, extra, partial), and partial matches,\textbf{ the framework can be fine-tuned to emphasize aspects critical to particular applications.}

\section{Prompts}
\label{sec:prompts}
\begin{figure*}[ht]
    \centering
    \includegraphics[width=1.0\textwidth]{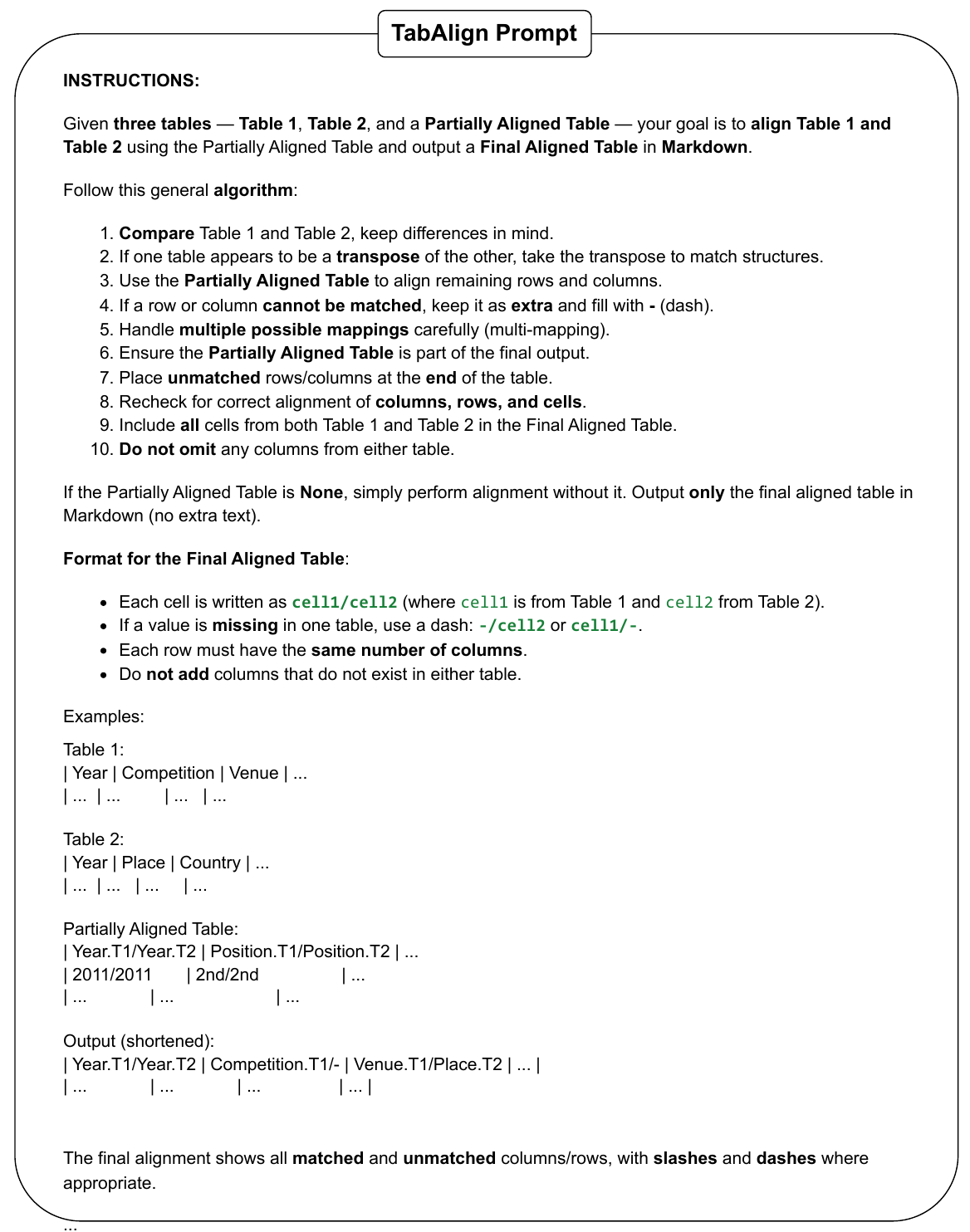}
    \caption{Prompt for tabular alignment, leveraging Partially Aligned Table and Reference Tables to generate a final structured table while preserving unmatched elements.}
    \label{fig:tabalign-prompt}
\end{figure*}

\begin{figure*}[ht]
    \centering
    \includegraphics[width=1.0\textwidth]{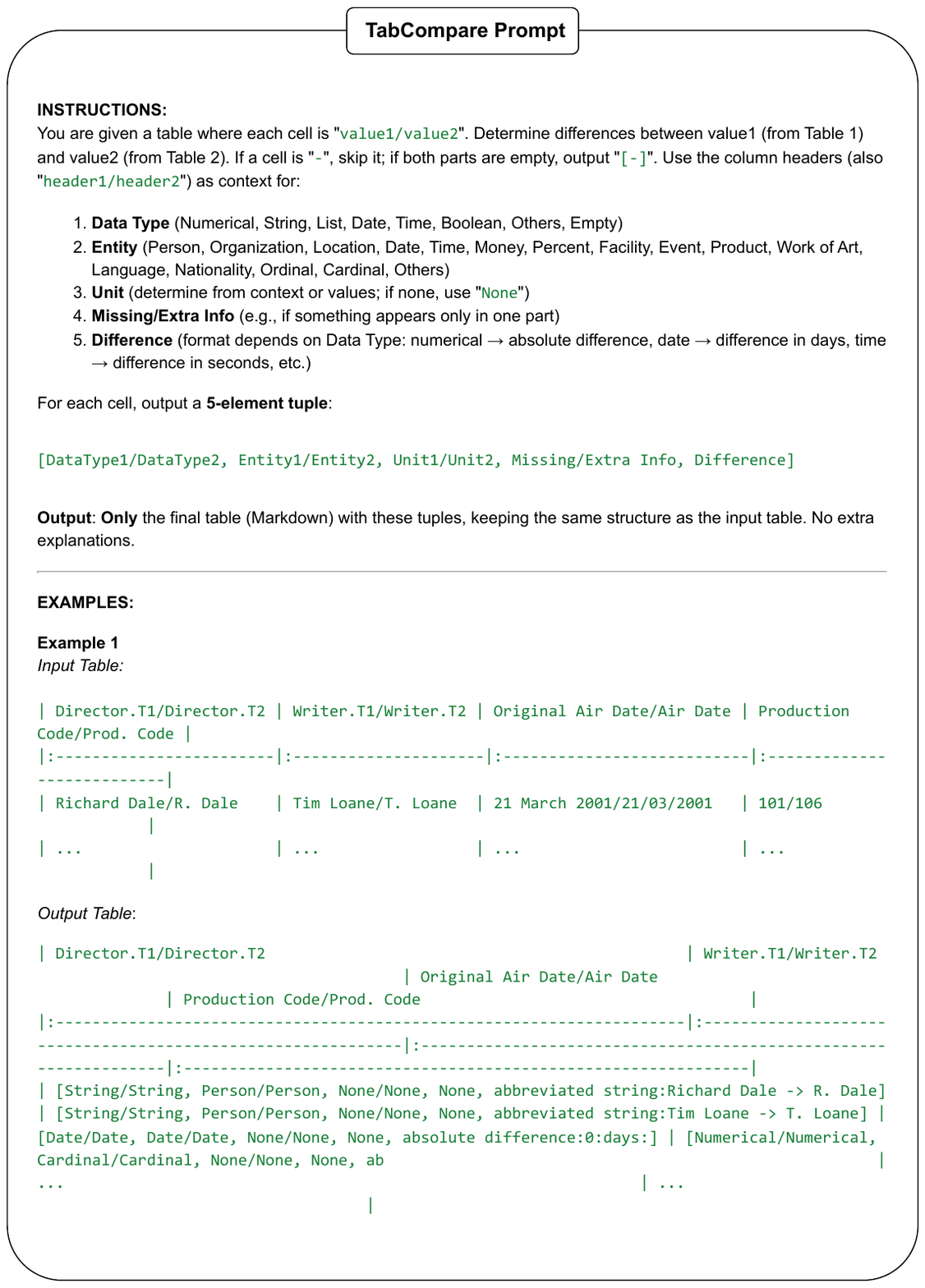}
    \caption{Prompt for identifying data type, entity, and unit differences between two tables, outputting structured tuples to capture variations in numerical, string, date, and categorical values.}
    \label{fig:tabcompare-prompt}
\end{figure*}

\begin{figure*}[ht]
    \centering
    \includegraphics[width=1.0\textwidth]{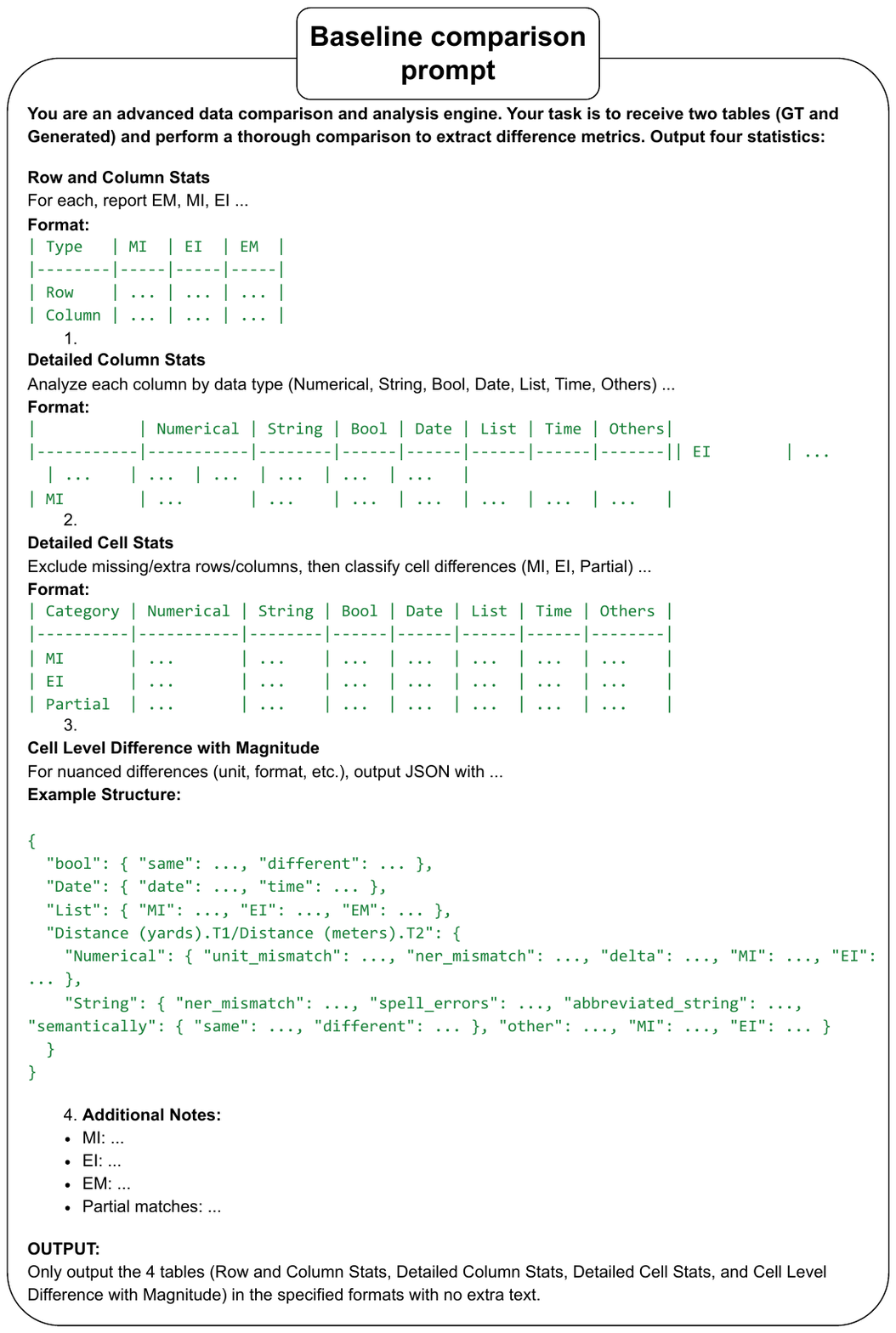}
    \caption{Baseline comparison prompt for evaluating differences between ground truth (GT) and generated data, providing structured metrics for row, column, and cell-level analysis.}
    \label{fig:baseline-prompt}
\end{figure*}

We provide detailed prompts for TabAlign in ~\autoref{fig:tabalign-prompt}, TabCompare in ~\autoref{fig:tabcompare-prompt} and Direct-LLM Baseline in ~\autoref{fig:baseline-prompt}.

\section{Qualitative Example}

\begin{figure*}[ht]
    \centering
    \includegraphics[width=0.95\textwidth]{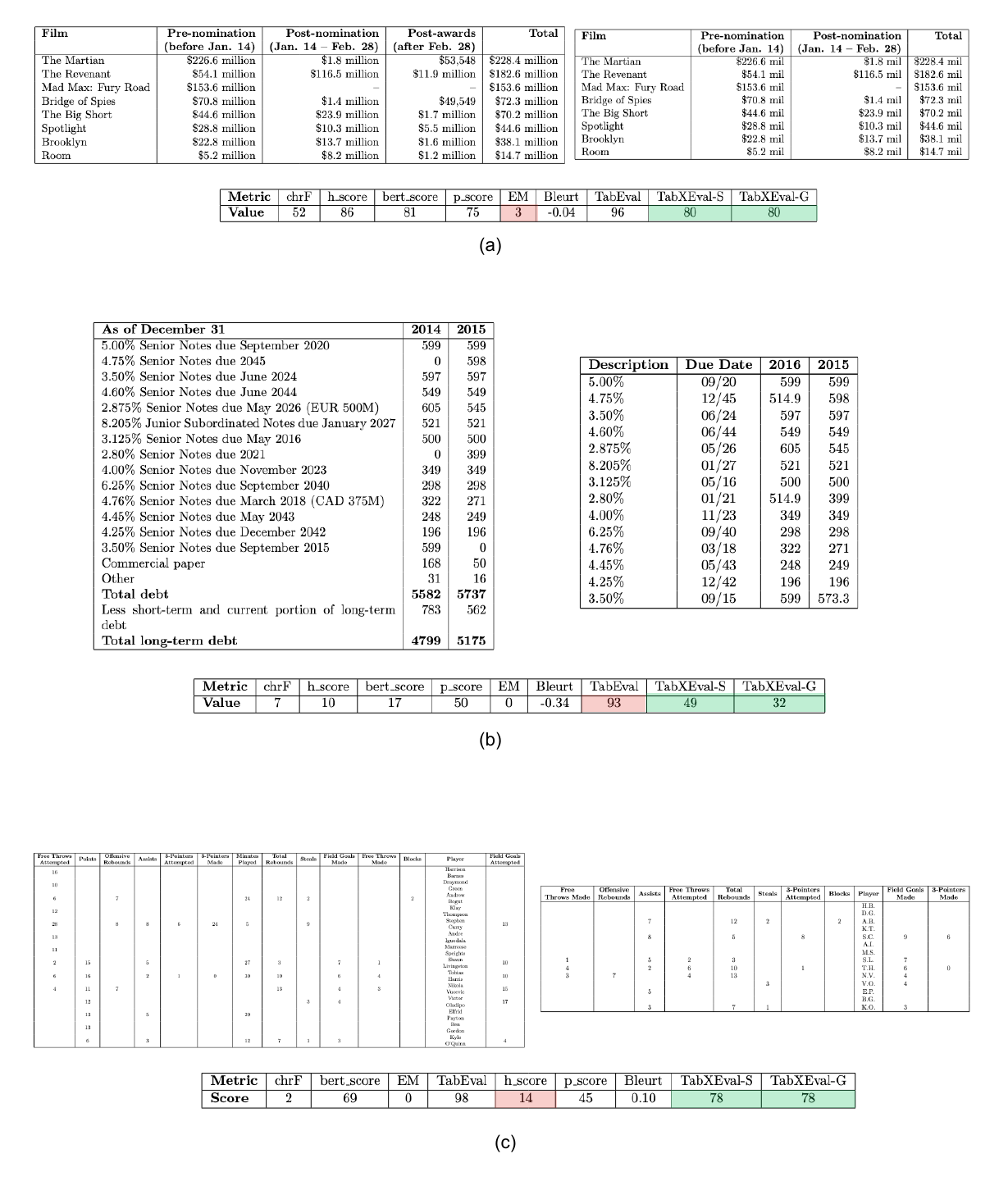}
\caption{Sample perturbed tables from the \tabbench{} benchmark, illustrating domain‐specific corruptions at different difficulty levels.  
  \textbf{(a) Movie domain with Easy:} ``Easy'' perturbations applied to a clean movie‐metadata table, including minor spelling errors in film titles, superficial header rephrasing, simple date‐format conversions (e.g., “March 3, 2020” $\leftrightarrow$ “03/03/2020”), trivial numeric formatting changes (addition/removal of thousands separators), and basic unit shifts (e.g., runtime in minutes vs.\ hours).  
  \textbf{(b) Finance domain with (Easy + Hard):} A financial report table subjected to both ``Easy'' (currency‐symbol normalization, decimal rounding) and ``Hard'' modifications, such as inconsistent metric abbreviations (e.g., “Rev.” vs.\ “Revenue”), merged indicator columns, omitted quarterly rows, and large‐scale unit mismatches (millions vs.\ billions).  
  \textbf{(c) Sports domain with Medium :} A sports‐stats table with ``Medium'' perturbations, featuring moderate header reordering (e.g., swapping “Team” and “Position”), slight numeric shifts in game statistics (win/loss counts adjusted by one or two), merged athlete performance rows, and partial row/column transpositions to emulate realistic table‐generation errors.}
    \label{fig:examples}
\end{figure*}

Figure~\ref{fig:examples} presents sample perturbed tables from the \tabbench{} benchmark, illustrating domain-specific corruptions across varying difficulty levels. It includes qualitative examples and a performance comparison of \textbf{\tabx{}} against prior metrics.

\label{sec:QE}

\section{Hierarchial Tables}
\label{sec:hierarchial_table}

Though direct processing of complex hierarchical tables as shown in table \ref{tab:hierarchial_table} (e.g., nested merged cells, multi-level headers) is currently a limitation, we've ideated a viable workaround. We can employ a preliminary structure decoding step, as illustrated in table \ref{tab:flat_table}, which effectively unrolls hierarchical headers into a flattened format. For example, parent header 'Release Date' and sub-headers 'Month'/'Year' transform into 'Release Date.Month' and 'Release Date.Year.' This standardized table representation is then fully compatible with our evaluation framework, accommodating both strict and relaxed mapping criteria

\begin{table}[htbp]
  \centering
  \caption{Hierarchical table}
  \begin{tabular}{llccr}
    \toprule
    \multicolumn{1}{c}{Movies} &%
    \multicolumn{1}{c}{Parts} &%
    \multicolumn{2}{c}{Release Date} &%
    \multicolumn{1}{c}{Ratings} \\
    \cmidrule(lr){3-4}
                                &       & Month & Year & \\ \midrule
    \multirow{2}{*}{When we die} & Part 1 & May       & 2010 & 5 \\
                                 & Part 2 & December  & 2022 & 3 \\ \bottomrule
  \end{tabular}
  \label{tab:hierarchial_table}
\end{table}

\begin{table}[htbp]
  \centering
  \caption{Flattened formatted table}
  \begin{tabular}{lccr}
    \toprule
    Movies $\vee$ Parts &
    Release Date $\wedge$ Month &
    Release Date $\wedge$ Year &
    Ratings \\ \midrule
    When we die $\vee$ Part 1 & May       & 2010 & 5 \\
    When we die $\vee$ Part 2 & December  & 2022 & 3 \\ \bottomrule
  \end{tabular}
  \label{tab:flat_table}
\end{table}

\clearpage

\section{Human Evaluation Setup}
\label{sec:Human-Eval}
\noindent
\includegraphics[
  page=1,
  width=\textwidth
]{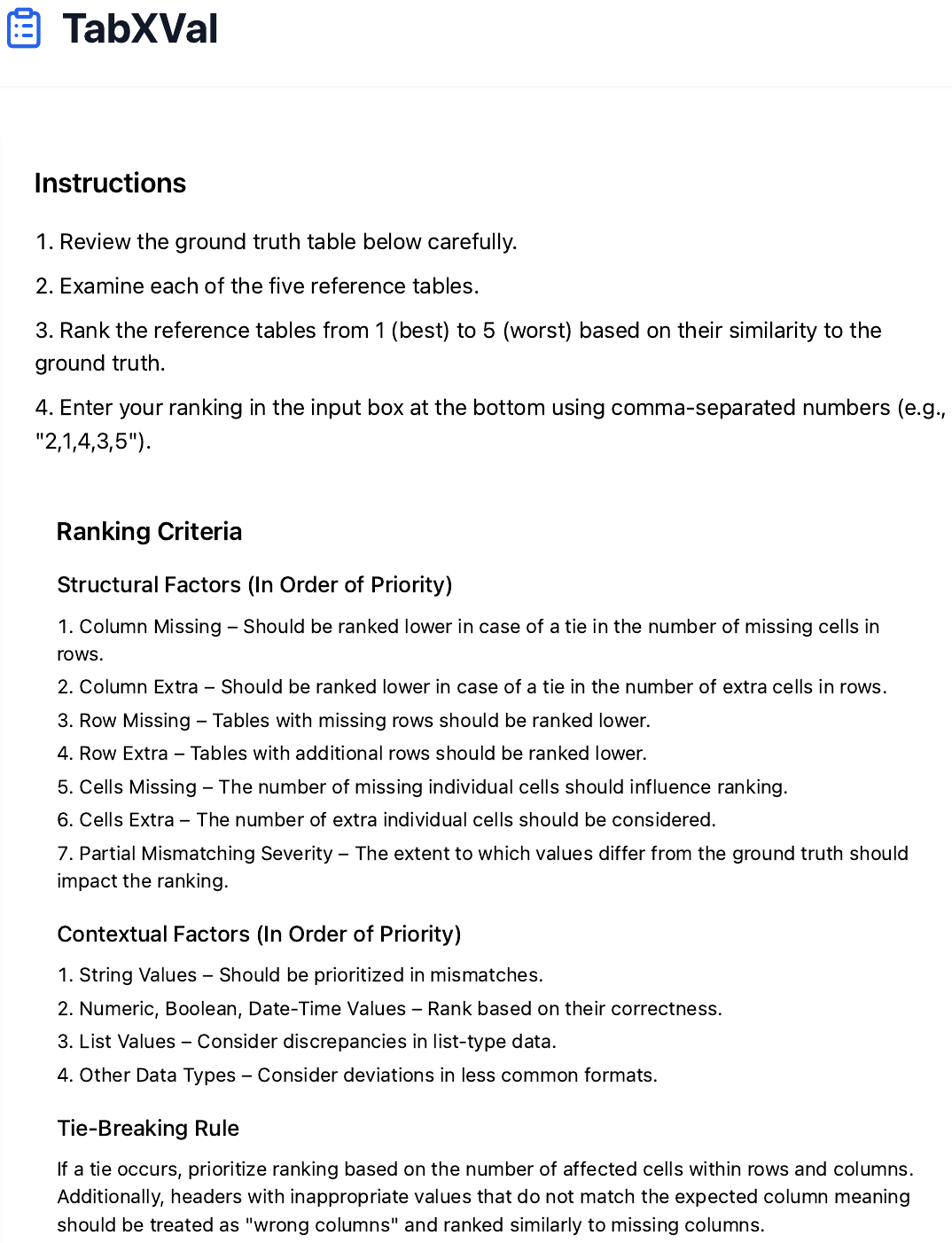}

\newpage
\noindent
\includegraphics[
  page=2,
  width=\textwidth
]{images/Human-ranking-web.pdf}

\newpage
\noindent
\includegraphics[
  page=3,
  width=\textwidth
]{images/Human-ranking-web.pdf}

\newpage
\noindent
\includegraphics[
  page=4,
  width=\textwidth
]{images/Human-ranking-web.pdf}

\end{document}